\documentclass[a4paper,conference]{IEEEtran}
\usepackage[dvipsnames]{xcolor}

\definecolor{mypink1}{rgb}{0.858, 0.188, 0.478}
\usepackage{dingbat}
\ifCLASSINFOpdf
  \usepackage[pdftex]{graphicx}
\else
  \usepackage[dvips]{graphicx}
\fi
\usepackage{amsmath}
\usepackage{bbding}
\usepackage{pifont}
\usepackage{wasysym}
\usepackage{amssymb}
\usepackage{url}

\begin{document}
%
\title{MFST: Multi-Features Siamese Tracker}

\author{\IEEEauthorblockN{Zhenxi Li, Guillaume-Alexandre Bilodeau}
\IEEEauthorblockA{LITIV Laboratory\\
Polytechnique Montr\'{e}al\\
Montr\'{e}al, Canada\\
zhenxi.li@polymtl.ca, gabilodeau@polymtl.ca}

\and
\IEEEauthorblockN{Wassim Bouachir}
\IEEEauthorblockA{Department of Science and Technology\\
T\'{E}LUQ University\\
Montr\'{e}al, Canada\\
wassim.bouachir@teluq.ca}}


%


\maketitle

\begin{abstract}
Siamese trackers have recently achieved interesting results due to their balance between accuracy and speed. This success is mainly due to the fact that deep similarity networks were specifically designed to address the image similarity problem. Therefore, they are inherently more appropriate than classical CNNs for the tracking task. However, Siamese trackers rely on the last convolutional layers for similarity analysis and target search, which restricts their performance. In this paper, we argue that using a single convolutional layer as feature representation is not the optimal choice within the deep similarity framework, as multiple convolutional layers provide several abstraction levels in characterizing an object. Starting from this motivation, we present the Multi-Features Siamese Tracker (MFST), a novel tracking algorithm exploiting several hierarchical feature maps for robust deep similarity tracking. MFST proceeds by fusing hierarchical features to ensure a richer and more efficient representation. Moreover, we handle appearance variation by calibrating deep features extracted from two different CNN models. Based on this advanced feature representation, our algorithm achieves high tracking accuracy, while outperforming several state-of-the-art trackers, including standard Siamese trackers. The code and trained models are available at \texttt{\color{mypink1} \url{https://github.com/zhenxili96/MFST}}.

\end{abstract}


%
\IEEEpeerreviewmaketitle

\section{Introduction}
\label{intro}
During the last few years, deep learning trackers achieved a stimulating effect by bringing new ideas to object tracking. This paradigm has become successful mainly due the use of convolutional neural network (CNN)-based features for appearance modeling. While several tracking methods used classification-based CNN models that are built following the principles of visual classification tasks, another approach \cite{SiamFC} formulates the tracking task as a deep similarity learning problem, where a Siamese network is trained to locate the target within a search image. This method uses feature representations extracted by CNNs and performs correlation operation with a sliding window to calculate a similarity map for target location. Rather than detecting by correlation, other deep similarity trackers \cite{SiamRPN,DaSiamRPN,GOTURN} generate the bounding box for the target object by regression networks. For example, GOTURN \cite{GOTURN} predicts the bounding box of the target object with a simple CNN model. The trackers \cite{DaSiamRPN} and \cite{SiamRPN} generate a number of proposals for the target after extracting feature representations. Classification and regression procedures are then applied to produce the final tracking.

By formulating object tracking as a deep similarity problem, Siamese trackers achieved significant progress in terms of both speed and accuracy. However, less efforts have been devoted to advance the feature representation power of these models. In fact, Siamese trackers typically rely on features from the last convolutional layers for similarity analysis and target state prediction. In this work, we argue that this is not the optimal choice, and demonstrate that features from earlier layers are important for improving tracking accuracy. Indeed, the combination of several convolutional layers was shown to be efficient for robust tracking \cite{HCFT,HCFTX}, as features from different layers provide different levels of information on the object. In particular, the last convolutional layers retain general characteristics represented in summarized fashion, while the first convolutional layers provide low-level features. These latter are extremely valuable for precise localization of the target, as they are more object-specific and capture finer spatial details. 

Therefore, instead of using features from a single CNN model, we propose to exploit different models within the deep similarity framework. Diversifying feature representations significantly improves tracking performance, and such an approach is shown to ensure robustness against appearance variation \cite{MBST}, one of the most challenging tracking difficulties.

Based on these principles, we propose the Multi-Features Siamese Tracker (MFST). MFST exploits diverse features from several convolutional layers and utilizes two models with proper feature fusing strategies to achieve an improved tracking performance.  Our contributions can be summarized as follows. Firstly, we explore feature fusing strategies with a feature recalibration module to make a better use of feature representations. Secondly, we exploit feature representations from several hierarchical convolutional layers as well as different models for object tracking. Thirdly, we present the MFST algorithm, that achieves an improved performance compared to recent state-of-the-art trackers.

\section{Related Work}\label{sec:mfst_related}
\subsection{Deep similarity tracking} 
Siamese trackers formulate the tracking task as a similarity learning problem. First, the deep similarity network is trained during an offline phase to learn a general similarity function. The model is then applied for online tracking to evaluate the similarity between two network inputs: the target template and the current frame. The pioneering work SiamFC \cite{SiamFC} applied two identical branches made up of fully convolutional neural networks to extract the feature representations, on which cross correlation is computed to generate the tracking result. SiamFC outperformed several state-of-art trackers, while achieving real-time speed. Several improvements were subsequently proposed. For example, rather than performing correlation on deep features directly, CFNet \cite{CFNet} trained a correlation filter based on the extracted features of object to speed-up tracking. SA-Siam \cite{SASiam} encoded the target by a semantic branch and an appearance branch to improve tracking robustness. But since these siamese trackers only use the output of the last convolutional layers, more detailed target specific information from earlier layers is not exploited. In our work, we propose a siamese tracker that combines features from different hierarchical levels.

\begin{figure*}[!t]
\centering
\includegraphics[width=4.5in]{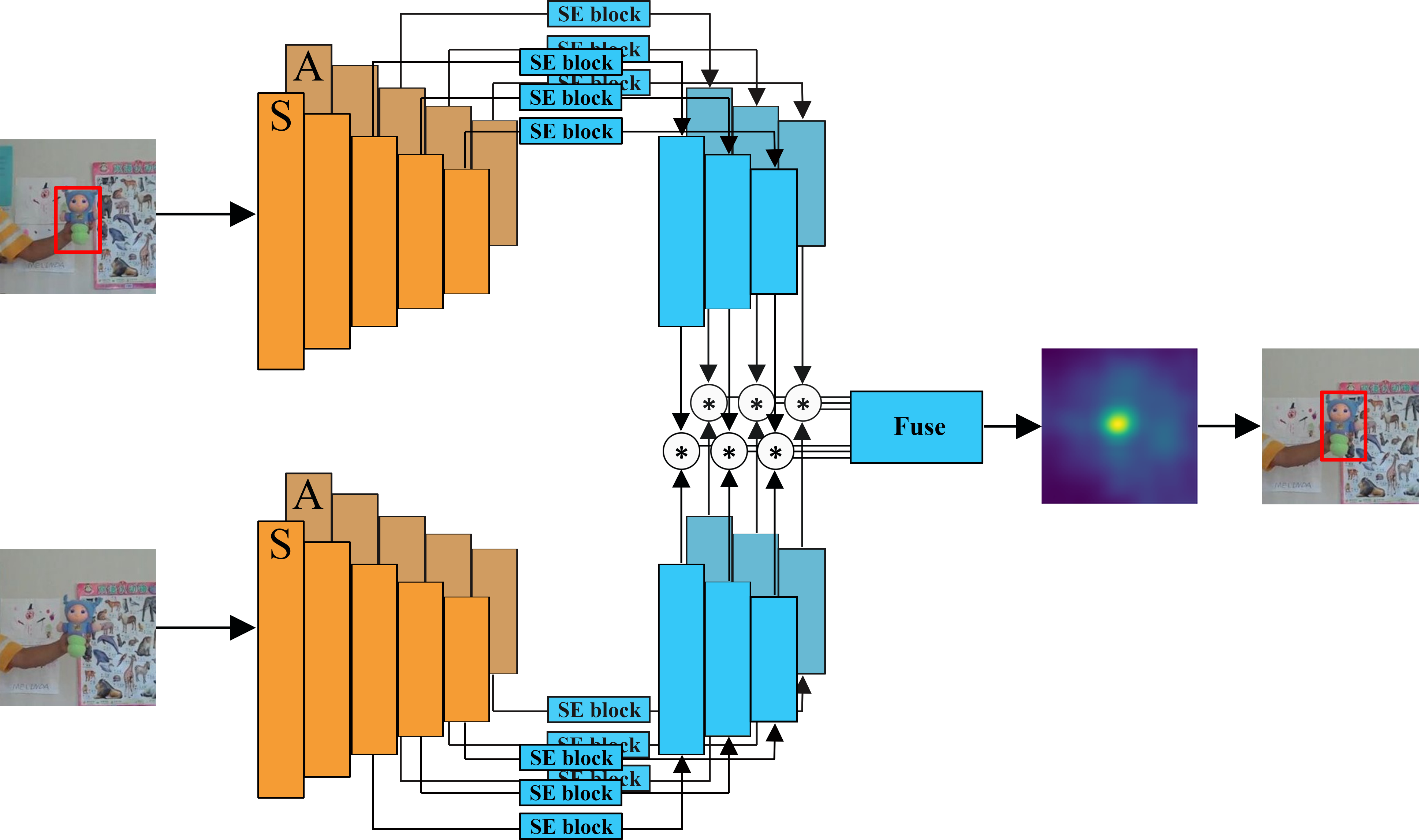}
\caption{The architecture of our MFST tracker. Two CNN models are utilized as feature extractors. Their features are calibrated by Squeeze-and-Excitation (SE) blocks. Correlations are then applied over the features of the search region with the features of the exemplar patch. The output response maps are fused to calculate the new position of the target. Bright orange: SiamFC (S) and dark orange: AlexNet (A).}
\label{fig:architecture}
\end{figure*}

\subsection{Exploiting multiple hierarchical levels in CNNs for tracking} 
Most CNN-based trackers used only the output of the last convolutional layer that contains semantic information represented in a summarized fashion. However, different convolutional layers embed different levels of visual abstraction. In fact, convolutional layers provide several detail levels in characterizing an object, and the combination of different levels is demonstrated to be efficient for robust tracking \cite{HCFT, li2018deep}. In this direction, the pioneering method HCFT \cite{HCFT} tracks the target using correlation filters learned on each layer. That work demonstrated that the combination of several hierarchical convolutional features allows to improve tracking robustness. Subsequently, \cite{HCFTX} presented a visualization of features extracted from different convolutional layers. In their work, they employed three convolutional layers as the target object representations, which are then convolved with the learned correlation filters to generate the response map, and a long-term memory filter to correct results. The use of multiple hierarchical features allowed their method to be much more robust.

\subsection{Multi-Branch Tracking} 
Target appearance variation is one of the most challenging problems in object tracking. In fact, the object appearance may change significantly during tracking due to several factors (e.g. deformation, 3D rotation). Thus, a single fixed network cannot guarantee discriminative feature representations in all tracking situations. To handle the problem of target appearance variation, TRACA \cite{TRACA} trained multiple auto-encoders, each for different appearance categories. These auto-encoders compress the feature representation for each category. The best expert auto-encoder is selected by a pretrained context-aware network. By selecting a specific auto-encoder for the tracked object, a more robust representation can be generated for the tracking. MDNet \cite{MDNet} applied a fixed CNN for feature extraction, but used multiple regression branches for objects belonging to different tracking scenarios. More recently, MBST \cite{MBST} extracted the feature representation for the target object through multiple branches and selected the best branch according to their response maps. With multiple branches MBST can obtain diverse feature representations and select the most discriminative one under the prevailing circumstance. In their study, we can observe that the greater the number of branches, the more robust the tracker is. However, This is achieved at the cost of higher computation time. In this work, we can get a diverse feature representation of a target at lower cost because some of the representations are extracted from many layers of the CNNs. Therefore, we do not need a large number of siamese branches.

\section{The proposed MFST tracker}\label{sec:mfst_main}
 In our method design, we consider that features from different convolutional layers provide different levels of abstraction, and that the different channels of the features play different roles in tracking. Therefore, we recalibrate deep features extracted from CNN models and combine hierarchical levels to ensure a more efficient representation. Besides, since models trained for different tasks can diversify the feature representation as well, we build our siamese architecture with two CNN models to achieve better performance.

\subsection{Overview of the network architecture}
The network architecture of our tracker is shown in Figure \ref{fig:architecture}. We use two different pretrained CNN models as feature extractors, SiamFC \cite{SiamFC} and AlexNet \cite{AlexNet}, as indicated in Figure \ref{fig:architecture}. The two models are denoted as $S$ and $A$ respectively in the following explanations. Both of them are five layers fully convolutional neural networks.

\begin{figure*}[t]
\centering
\includegraphics[width=2.5in]{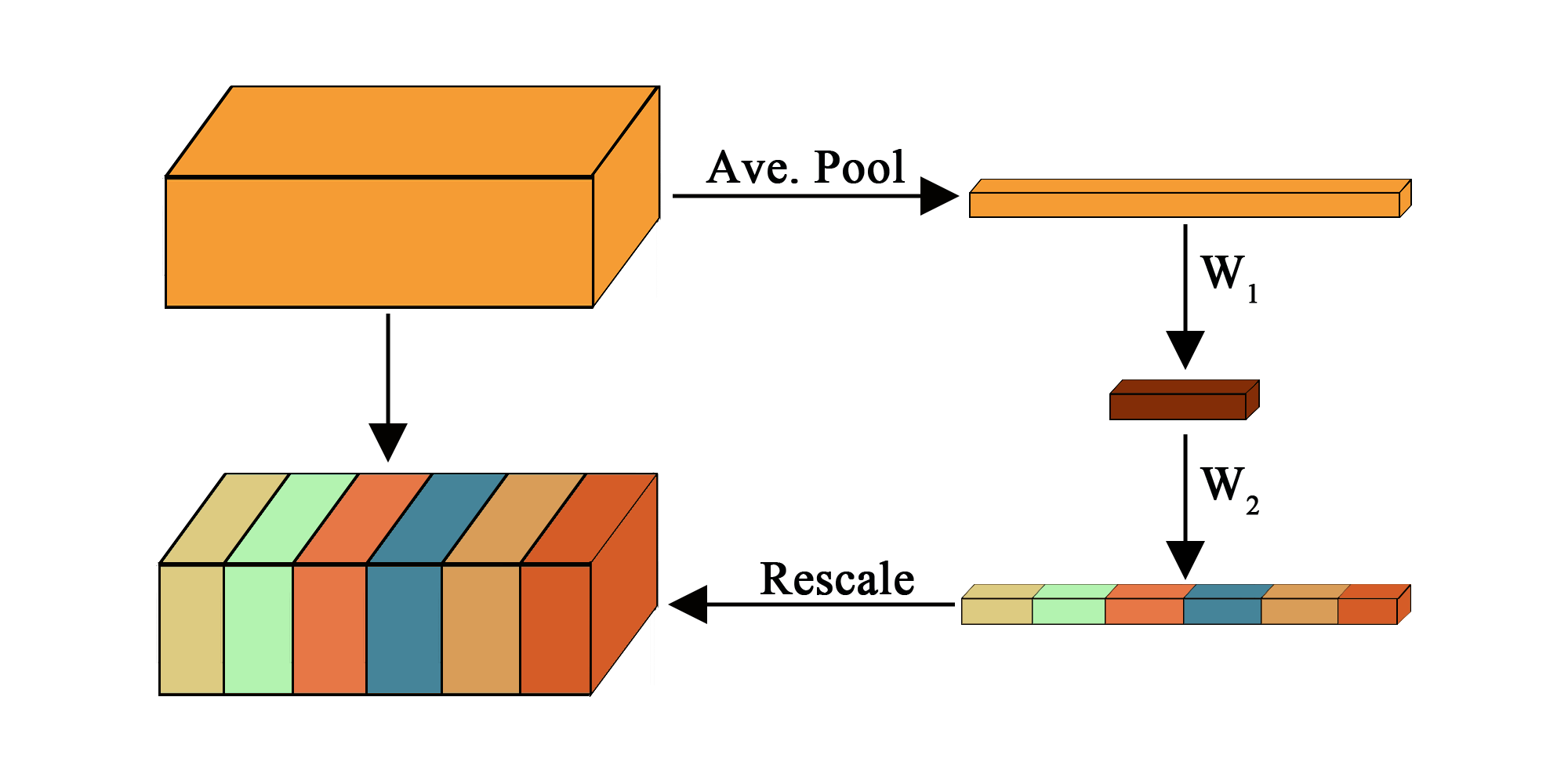}
\caption{Illustration of the SE-block. The SE operation comprises two steps: squeeze and excitation. The squeeze step uses average pooling operation to generate the channel descriptor. The excitation step uses a two layers MLP to capture channel-wise dependencies.}
\label{fig:seblock}
\end{figure*}

The input of our method consists of an exemplar patch $z$ cropped according to the initial bounding box or the result of last frame and search region $x$. The exemplar patch has a size of $W_{z}\times H_{z}\times 3$ and the search region has a size of $W_{x}\times H_{x}\times 3$ ($W_{z}<W_{x}$ and $H_{z}<H_{x}$), representing the width, height and the color channels of the image patches. Since our method formulates the tracking task as a similarity learning problem, let $x$ be considered as a collection of candidate patches. 

With the two CNN models, we obtain the deep features $S_{l_{i}}$, $A_{l_{i}}$ ($l=c3,c4,c5$, $i=z,x$) from conv3, conv4, conv5 layers of each model. These are the preliminary deep feature representations of the inputs. All the features are then recalibrated with Squeeze-and-Excitation blocks (SE-blocks) \cite{SENet}. The recalibrated features are denoted as $S_{l_{i}}^{*}$, $A_{l_{i}}^{*}$, respectively for the two models. The details of SE-blocks are illustrated in Fig. \ref{fig:seblock}. The blocks are trained to explore the importance of the different channels for tracking. They learn weights for different channels to recalibrate features extracted from the preliminary models. 

Once feature representations are generated, we apply cross-correlation operations for each recalibrated feature map pairs to generate response maps. The cross-correlation operation can be implemented by a convolution layer using the feature of the exemplar as filter. Then, we fuse these response maps to produce the final response map. The new target position is selected as the maximum value location in the final response map.

\subsection{Feature extraction}

\paragraph{Hierarchical convolutional features} It is well known that the last convolutional layers of CNNs encode more semantic information, which is invariant to significant appearance variations compared to earlier layers. However, its resolution is too coarse (due to the large receptive field) for precise localization. On the contrary, features from earlier layers contain less semantic information, but they retain more spatial details and are more precise in localization. Thus, we propose to exploit multiple hierarchical levels of features to build a better representation of the target. 

We use the convolutional layers of two lightweight pretrained CNN models as feature extractors: SiamFC \cite{SiamFC} and AlexNet \cite{AlexNet}. The two models are trained for object tracking and image classification tasks, respectively. 
We use features extracted from the 3rd, 4th, 5th convolutional layers as the preliminary target representations.

\paragraph{Feature recalibration} Considering that different channels of deep features play different roles in tracking, we apply SE-blocks \cite{SENet} over the raw deep features extracted from the basic models. An illustration of SE-block is shown in Fig. \ref{fig:seblock}. The SE-block includes two operations: 1) squeeze and 2) excitation. The squeeze step corresponds to an average pooling operation. Given a 3D feature map, this operation generates the channel descriptor $\mathbf{\omega}_{sq}$ as follows:
\begin{equation}
    \mathbf{\omega}_{sq}=\frac{1}{W\times H}\sum_{m=1}^{W}\sum_{n=1}^{H}v_{c}(m,n), (c=1,...,C)
\end{equation}
where $W$, $H$, $C$ are the width, height and the number of channels of the deep feature, and $v_{c}(m,n)$ is the corresponding value in the feature map. The subsequent step is the Excitation through a two layers Multi-layer perceptron (MLP). Its goal is to capture the channel-wise dependencies that can be expressed as:
\begin{equation}
    \mathbf{\omega}_{ex}=\sigma(\mathbf{W_{2}}\delta(\mathbf{W_{1}}\mathbf{\omega}_{sq}))
\end{equation}
where $\sigma$ is a sigmoid activation, $\delta$ is a ReLU activation, $\mathbf{W}_{1}\in R^{\frac{C}{r}\times C}$ and $\mathbf{W}_{2}\in R^{C\times\frac{C}{r}}$ are the weights for each layer, and $\mathbf{r}$ is the channel reduction factor used to change the dimension. After the excitation operation, we obtain the channel weight $\omega_{ex}$. The weight is used to rescale the feature maps extracted by the basic models:
\begin{equation}
    F_{l_{i}}^{*}=\mathbf{\omega}_{ex}\cdot F_{l_{i}},
\end{equation}
where $\cdot$ is a channel-wise multiplication and $F=(S,A)$. Note that $\omega_{ex}$ is learned for each layer in a basic model, but the corresponding layers for the CNNs of the exemplar patch and the search region use the same channel weight. We train the SE-blocks to obtain six $\omega_{ex}$ in total.

\subsection{Combining response maps}
Once the feature representations from convolutional layers of each model are obtained, we apply cross-correlation operation, which is implemented by convolution over the corresponding feature maps to generate the response map $r$ as:
\begin{equation}
    r(z,x)=corr(F^{*}(z),F^{*}(x)),
\end{equation}
where $F^{*}$ is the weighted feature map generated by the CNN model and SE-block. The response maps are then combined. For a pair of image input, six response maps are generated. They are denoted as $r_{c3}^{S}$, $r_{c4}^{S}$, $r_{c5}^{S}$, $r_{c3}^{A}$, $r_{c4}^{A}$ and $r_{c5}^{A}$. Note that we do not need to rescale the response maps for combination, since they have the same size (see Section \ref{sec:detail}, Data Dimensions).

The response maps are combined hierarchically. After fusing $r^{S}$ and $r^{A}$ for the two CNN models, we combine the two response maps to get the final map. The combination is performed by considering three strategies: hard weight (HW), soft mean (SM) and soft weight (SM) \cite{HCFTX}, illustrated as follows:
\begin{equation}
    \textrm{Hard weight: } r^{*}=\sum_{t=1}^{N}w_{t}r_{t}
\end{equation}
\begin{equation}
    \textrm{Soft mean: } r^{*}=\sum_{t=1}^{N}\frac{r_{t}}{max(r_{t})}
\end{equation}
\begin{equation}
    \textrm{Soft weight: } r^{*}=\sum_{t=1}^{N}\frac{w_{t}r_{t}}{max(r_{t})}
\end{equation}
where $r^{*}$ is the combined response map, $N$ is the number of response maps to be combined together, and $w_{t}$ is the empirical weight for each response map.

The optimal weights of HW and SW are obtained by experiments. We choose the corresponding strategy that achieves the best performance on the OTB benchmarks to generate the combined response maps from each model, which are denoted as $r^{S}$ and $r^{A}$. As illustrated in Table \ref{tab:table2}, we then test the three different strategies again to find the best strategy to combine $r^{S}$ and $r^{A}$. Finally, the corresponding location of the maximum value in the final response map is selected as the new location of the target. 

\begin{table*}[!t]
    \centering
    \caption{Experiments with several variations of our method, where A and S denote respectively the AlexNet model and SiamFC model. \textbf{Boldface} indicates best results. }
    \begin{tabular}{|c|c c c c c|c c|c c|c c|}
    \hline
       & & & & & &
      \multicolumn{2}{c|}{OTB-2013} &
      \multicolumn{2}{c|}{OTB-50} &
      \multicolumn{2}{c|}{OTB-100}\\
        Model& Conv3 & Conv4 & Conv5 & Fusion & SE &  AUC & Prec. & AUC & Prec. & AUC & Prec. \\
        \hline
        A&\checkmark & & & & & 0.587&0.740 & 0.474& 0.618& 0.559& 0.712\\
        A&\checkmark & & & & \checkmark & 0.603 & 0.755 & 0.504 & 0.642 & 0.587 & 0.747\\
        A& & \checkmark & & & & 0.632 & 0.789& 0.536 & 0.692 & 0.614 & 0.778 \\
        A& & \checkmark & & & \checkmark & \textbf{0.637} & 0.801 & 0.544 & 0.707 & 0.623 & 0.795 \\
        A& & & \checkmark & & & 0.582&0.763 & 0.496& 0.665&0.557 & 0.735\\
        A& & & \checkmark & &\checkmark& 0.573 & 0.762 & 0.507 & 0.696 & 0.575 & 0.769\\
        A& \checkmark & \checkmark & \checkmark & HW & & 0.623 & 0.774 & 0.515 & 0.657 & 0.605 & 0.763 \\
        A& \checkmark & \checkmark & \checkmark & SM & & 0.633 & 0.797& 0.542 & 0.705 & 0.616 & 0.784\\
        A& \checkmark & \checkmark & \checkmark & SW & & 0.630 & 0.795 & 0.538 & 0.699 & 0.616 & 0.786 \\
        A& \checkmark & \checkmark & \checkmark & HW & \checkmark & 0.627 & 0.798 & 0.537 & 0.700 & 0.617 & 0.790 \\
        A& \checkmark & \checkmark & \checkmark & SM & \checkmark & 0.631 & 0.799 & 0.542 & 0.706 & 0.621 & 0.792 \\
        A& \checkmark & \checkmark & \checkmark & SW & \checkmark & 0.635 & \textbf{0.811} & \textbf{0.545} & \textbf{0.716} & \textbf{0.627} & \textbf{0.803} \\
         \hline
        \hline
        S& \checkmark & & & & & 0.510 & 0.661 & 0.439 & 0.574 & 0.512 & 0.656 \\
        S&\checkmark & & & & \checkmark & 0.545 & 0.709 & 0.465 & 0.608 & 0.532 & 0.687 \\
        S& & \checkmark & & & & 0.584 & 0.757 & 0.507 & 0.666 & 0.570 & 0.742\\
        S& & \checkmark & & & \checkmark & 0.592 & 0.772 & 0.518 & 0.686 & 0.581 & 0.758 \\
        S& & & \checkmark & & & 0.600 & 0.791 & 0.519 & 0.698 & 0.586 & 0.766 \\
        S& & & \checkmark & & \checkmark & 0.606 & 0.801 & 0.535 & \textbf{0.722} & 0.588 & 0.777 \\
        S& \checkmark & \checkmark & \checkmark & HW & & 0.614 & 0.794 & 0.532 & 0.692 & 0.602 & 0.776 \\
        S& \checkmark & \checkmark & \checkmark & SM & & 0.612 & 0.787 & 0.539 & 0.697 & 0.607 & 0.777 \\
        S& \checkmark & \checkmark & \checkmark & SW & & 0.615 & 0.808 & 0.534 & 0.705 & 0.600 & 0.780 \\
        S& \checkmark & \checkmark & \checkmark & HW & \checkmark & \textbf{0.627} & \textbf{0.823} & \textbf{0.542} & 0.716 & \textbf{0.606} & \textbf{0.787} \\
        S& \checkmark & \checkmark & \checkmark & SM & \checkmark & 0.591 & 0.761 & 0.501 & 0.649 & 0.575 & 0.736 \\
        S& \checkmark & \checkmark & \checkmark & SW & \checkmark & 0.603 & 0.780 & 0.518 & 0.673 & 0.590 & 0.759 \\
        \hline
    \end{tabular}
    \label{tab:table1}
\end{table*}

\section{Experiments}\label{sec:mfst_exp}
We firstly perform an ablation study by investigating the contribution of each module, in order to find the best response map combination strategy corresponding to optimal representations. Secondly, we present a performance comparison with recent state-of-the-art trackers.

\subsection{Implementation Details}\label{sec:detail}
\paragraph{Network Structure} We use SiamFC \cite{SiamFC} and AlexNet \cite{AlexNet} as feature extractors. The SiamFC network is a fully convolutional neural network (including five layers) trained on a video dataset for object tracking. The AlexNet network consists of five convolutional layers and three fully connected layers trained on an image classification dataset. We slightly modified the stride of AlexNet to obtain the same dimensions for the outputs of both CNN models. Since only deep features are needed to represent the target, we remove the fully connected layers of AlexNet and keep only the convolutional layers to extract features. 

\paragraph{Data Dimensions} The inputs of our method consist in the exemplar patch $z$ and the search region $x$. The size of $z$ is $127\times 127$ and the size of $x$ is $255\times 255$. The output feature maps of $z$ have sizes of $10\times 10\times 384$, $8\times 8\times 384$ and $6\times 6\times 256$ respectively. The output feature maps of $x$ have sizes of $26\times 26\times 384$, $24\times 24\times 384$, and $22\times 22\times 256$ respectively. Taking the features of $z$ as filters to perform a convolution on the features of $x$, the size of the output response maps are all the same, $17\times17$. The final response map is resized to the size of the input to locate the target.

\paragraph{Training} The SiamFC model is trained on the ImageNet dataset (\cite{ImageNet}) by considering only color images. The ImageNet dataset contains more that 4,000 sequences, about 1.3 million frames and 2 million tracked objects with ground truth bounding boxes. For the input, we take a pair of images and crop the exemplar patch $z$ in the center and the search region $x$ in another image. The SiamFC model is trained for 50 epochs with an initial learning rate of 0.01. The learning rate decays with a factor of 0.86 after each epoch. The AlexNet model is pretrained on the ImageNet dataset for the image classification task. We trained the SE-blocks for the two models separately in the same manner. For each model, the original parameters are fixed. We then apply SE-blocks on the output of each layer and take the recalibrated output of each layer as the output feature to generate the result for training. The SE-blocks are trained on the ImageNet dataset with 50 epochs with an initial learning rate of 0.01. The learning rate decays with a factor of 0.86 after each epoch. 

We performed our experiments on a PC with an Intel i7-3770 3.40 GHz CPU and a Nvidia Titan Xp GPU. The benchmark results are calculated by the Python implementation of the OTB toolkit (\cite{otb100}). The average testing speed of our tracker is 39 fps.

\paragraph{Tracking} We initialize our tracker with the coordinates of the bounding box in the first frame. The exemplar patch is fed into the basic SiamFC model and AlexNet model to generate the preliminary feature representations $S_{l_{z}}$, $A_{l_{z}} (l=c3, c4, c5)$. Then, SE-block is applied to produce the recalibrated feature maps $S^{*}_{l_{z}}$, $A^{*}_{l_{z}}$, which are used to produce response maps for tracking the target on all subsequent frames.

When processing the current frame, the tracker crops the region centered on the last object center position, generates the feature representations, and output the response maps by a correlation operation with the feature maps of the target object. The new position of the target is indicated by the the maximum value in the final combined response map.

\paragraph{Hyperparameters} The channel reduction factor $r$ in SE-blocks is set to 4. The empirical weights $w_{t}$s for $r_{c3}^{S}$, $r_{c4}^{S}$, $r_{c5}^{S}$, $r_{c3}^{A}$, $r_{c4}^{A}$ and $r_{c5}^{A}$ are fixed to 0.1, 0.3, 0.7, and 0.1, 0.6, 0.3, respectively. The empirical weights $w_{t}$s for $r^{S}$ and $r^{A}$ are 0.3 and 0.7. To handle scale variations, we search the target object over three scales $1.025^{\{-1, 0, 1\}}$ during evaluation and testing.

\subsection{Datasets and Evaluation Metrics}
 We evaluate our method on the OTB benchmarks \cite{otb50, otb100}, which consist of three datasets, OTB50, OTB2013 and OTB100. They contain 50, 51, and 100 video sequences. The benchmarks propose two evaluation metrics for quantitative analysis: (1) the center location error (CLE) and (2) the overlap score, which are used to produce precision and success plots respectively. The precision plot is obtained by calculating the ratios of successful tracking iterations at several CLE thresholds. The threshold of 20 pixels is used to rank the results. For the success plot, we compute the IoU (intersection over union) between the tracking results and the ground truth labels for each frame. The plot show the corresponding success rate for each overlap threshold. The AUC (area-under-curve) is used to rank the results.

\begin{table*}[!t]
    \centering
    \caption{Experiments on combining response maps from the two CNN models. $A_{conv5}$ is only taking features from the last convolutional layer of AlexNet network, $S_{conv5}$ is only taking features from the last convolutional layer of SiamFC network.  $A_{com}$ is the combined response map from AlexNet network by soft weight combination, $S_{com}$ is the combined response map from SiamFC network by hard weight combination. \textbf{Boldface} indicates best results. }
    \begin{tabular}{|c c c c|c c|c c|c c|}
    \hline
        & & & &
      \multicolumn{2}{c|}{OTB-2013} &
      \multicolumn{2}{c|}{OTB-50} &
      \multicolumn{2}{c|}{OTB-100}\\
        $A$ & $S$ & Fusion & SE &  AUC & Prec. & AUC & Prec. & AUC & Prec. \\
    \hline
    $A_{conv5}$ & & & &0.582&0.763&0.496&0.665&0.557&0.735\\
    $A_{com}$ & & & & 0.630 & 0.795 & 0.538 & 0.699 & 0.616 & 0.786\\
    $A_{com}$ & & &\checkmark & 0.635 & 0.811 & 0.545 & 0.716 & 0.627 & 0.803\\
    & $S_{conv5}$ & & &0.661&0.854&0.581&0.764&0.647&0.831\\
    & $S_{com}$ & & & 0.614 & 0.794 & 0.532 & 0.692 & 0.602 & 0.776 \\
    & $S_{com}$ & &\checkmark & 0.627 & 0.823 & 0.542 & 0.716 & 0.606 & 0.787\\
    $A_{com}$ & $S_{com}$ & HW & &0.637&0.815&0.555&0.720&0.625&0.801\\
    $A_{com}$ & $S_{com}$ & SM & &0.647&0.819&0.560&0.728&0.638&0.816\\
    $A_{com}$ & $S_{com}$ & SW & &0.647&0.818&0.564&0.734&0.637&0.813\\
    $A_{com}$ & $S_{com}$ & HW & \checkmark&0.667&0.852&\textbf{0.583}&0.761&0.644&0.824\\
    $A_{com}$ & $S_{com}$ & SM & \checkmark & 0.640 & 0.810 & 0.557 & 0.718 & 0.632 & 0.804\\
    $A_{com}$ & $S_{com}$ & SW & \checkmark &\textbf{0.667}&\textbf{0.854}&0.581&\textbf{0.764}&\textbf{0.647}&\textbf{0.831}\\
    \hline
    \end{tabular}
    \label{tab:table2}
\end{table*}

\begin{figure*}[t]
\centering
\includegraphics[width=2.5in]{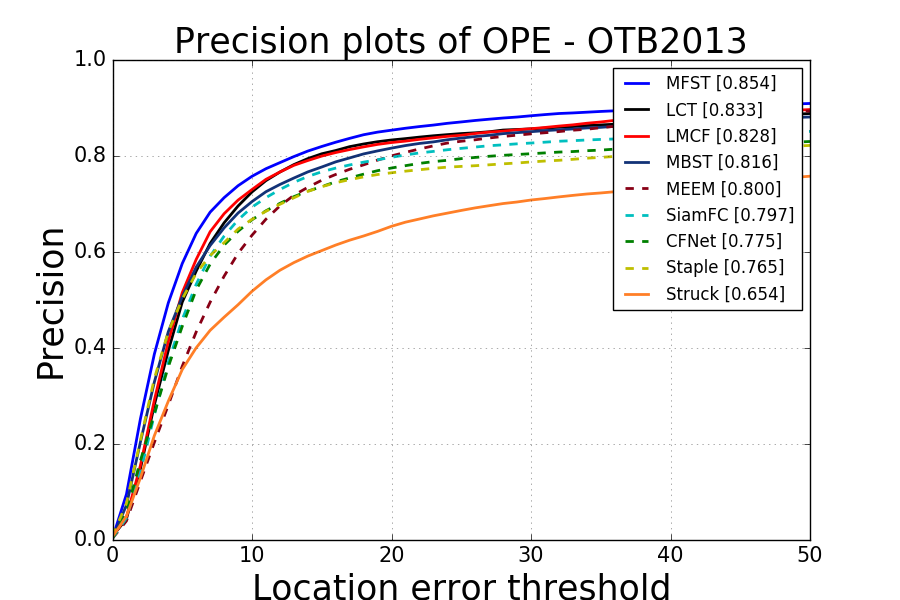}
\includegraphics[width=2.5in]{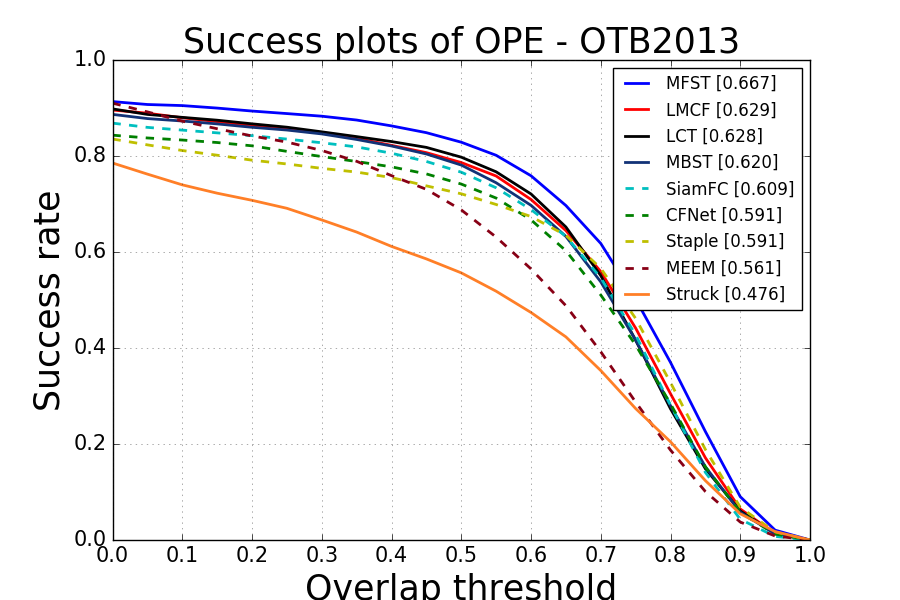}
{\includegraphics[width=2.5in]{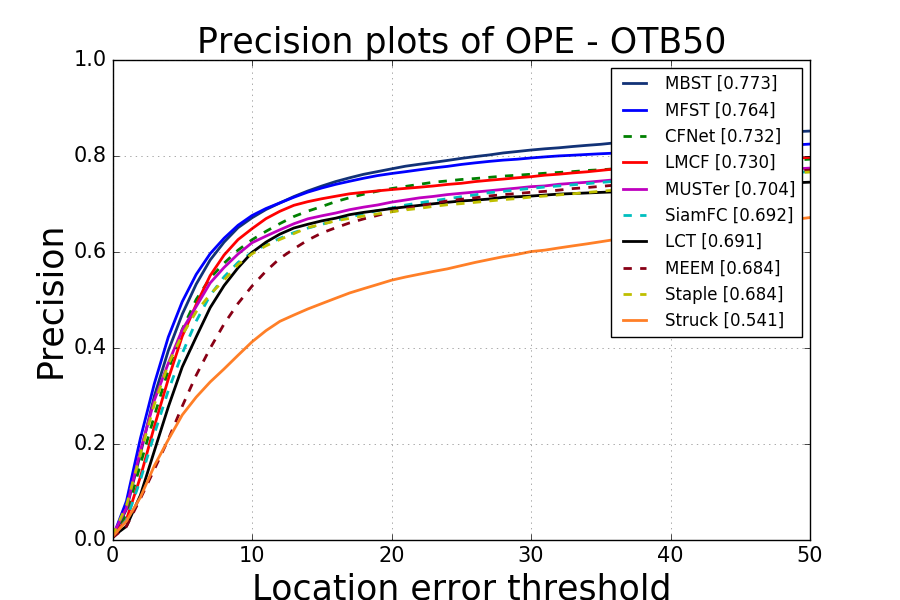}
\includegraphics[width=2.5in]{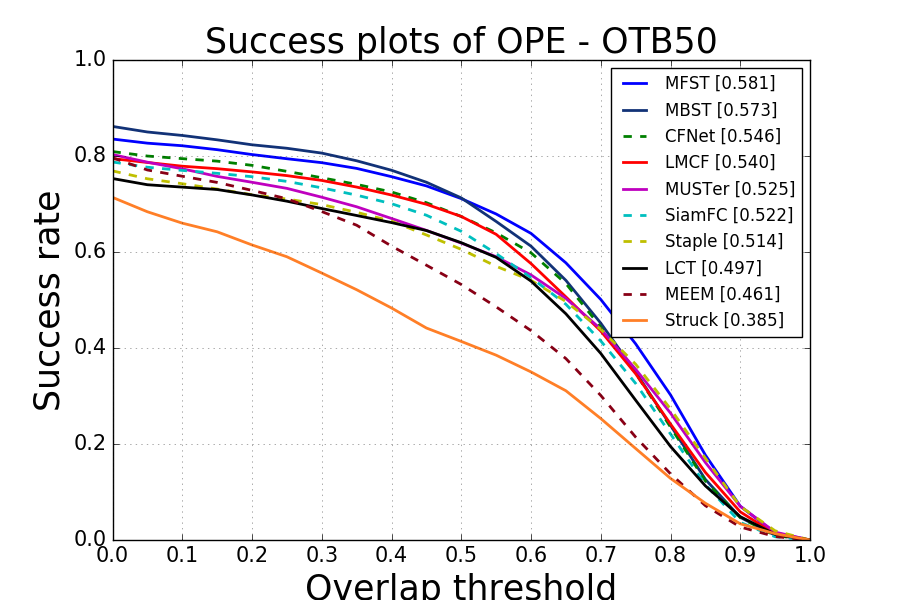}}
\includegraphics[width=2.5in]{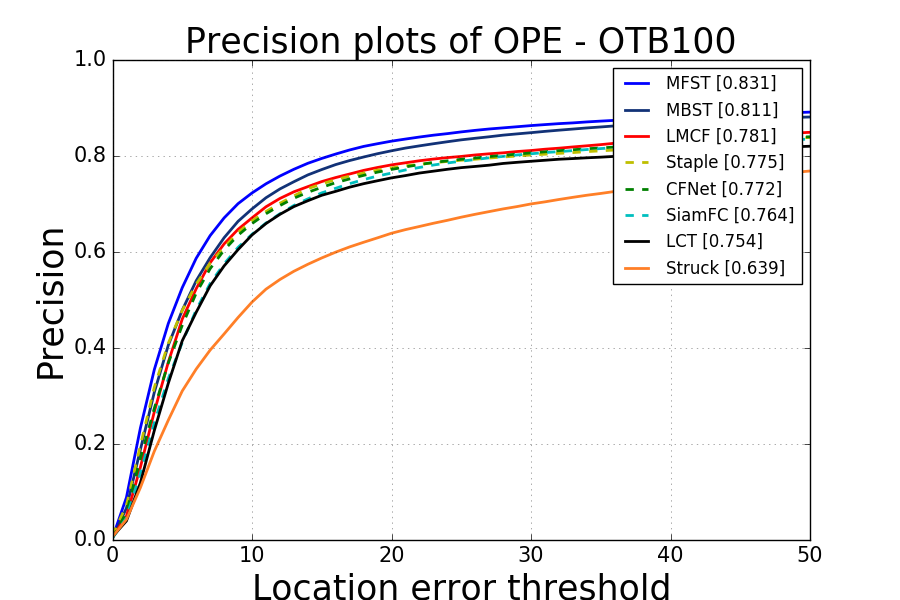}
\includegraphics[width=2.5in]{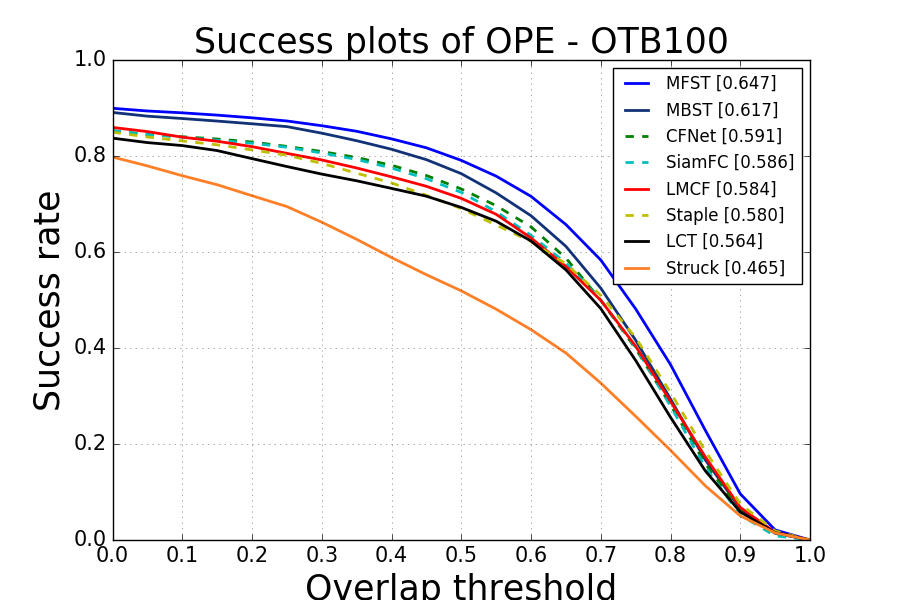}
\caption{The evaluation results on OTB benchmarks.}
\label{fig:cfst_comp}
\end{figure*}

\subsection{Ablation Analysis}
To investigate the contributions of each module and the optimal strategies to combine representations, we perform an ablation analysis with several variations of our method.

\paragraph{A proper combination of features is better than features from a single layer} As illustrated in Table \ref{tab:table1}, we experimented using features from a single layer as the target representation and combined features from several layers with different combination strategies for the two CNN models. The results show that, taken separately, conv3, conv4, conv5 produce similar results. Since object appearance changes, conv3 that should give the most precise location does not always achieve good performance. However, with a proper combination, the representation is significantly improved.

\paragraph{Features get enhanced with recalibration} Due to the Squeeze-and-Excitation operations, recalibrated features achieve better performance than the preliminary features. Recalibration through SE-block thus improves the representation power of features from single layer, which results in a better representation of the combined features.

\paragraph{Multiple models are better than a single model} Our approach utilizes two CNN models as feature extractors. Here we also conducted experiments to verify the benefit of using two CNN models. As illustrated in Table \ref{tab:table2}, we evaluated the performance of using one CNN model and using the combination of the two CNN models. The results show that the combination of two models is more discriminative than only one model regardless of the use SE-blocks.

\paragraph{A proper strategy is important for the response maps combination} We applied three strategies to combine the response maps: hard weight (HW), soft mean (SM) and soft weight (SW). Since the two CNN models we used are trained for different tasks and that features from different layers embed different level of information, different types of combination strategies should be applied to make the best use of features. The experimental results show that generally, combined features are more discriminative than independent features, while a proper strategy can improve the performance significantly as illustrated in Table \ref{tab:table1} and Table \ref{tab:table2}. In addition, we observe that the soft weight strategy is generally the most appropriate, except for combining hierarchical features from the SiamFC model.

\begin{table*}[!t]
    \centering
    \caption{Speed evaluation results on OTB benchmarks.}
    \begin{tabular}{|c|c|c|c|c|c|c|c|c|c|c|}
    \hline
        \textbf{Tracker} & MFST & MBST & LCT & LMCF & MEEM & CFNet & SiamFC & Staple & Struck & MUSTer \\
        \hline
        
        \textbf{Speed (fps)} & 39 & 17 & 27 & 66 & 22 & 75 & 86 & 41 & 10 & 5
        \\
    \hline
    \end{tabular}
    \label{tab:speed_table}
\end{table*}

\subsection{Comparison with state-of-the-art trackers}
We compare our tracker MFST with MBST \cite{MBST}, LMCF \cite{LMCF}, CFNet \cite{CFNet}, SiamFC \cite{SiamFC},  Staple \cite{Staple}, Struck \cite{Struck}, MUSTER \cite{MUSTer}, LCT \cite{LCT}, MEEM \cite{MEEM} on OTB benchmarks \cite{otb50,otb100}. The precision and success plots are shown in Figure \ref{fig:cfst_comp}. Both plots show that our tracker MFST achieves the best performance among these state-of-the-art trackers on OTB benchmarks, except on the precision plot of OTB-50. The feature calibration mechanism we employed is beneficial for tracking as well. It demonstrates that by using the combined features, the target representation of our method is more efficient and robust. From the results, although we use siamese networks to address the tracking problem as SiamFC, and take SiamFC as one of our feature extractors, our tracker achieves much improved performance over SiamFC. Besides, despite the fact that the MBST tracker employs diverse feature representations from many CNN models, our tracker achieves better results with only two CNN models, in terms of both tracking accuracy and speed. A speed comparison is shown in Table \ref{tab:speed_table}.

\section{Conclusion}\label{sec:mfst_conclude}
In this paper, we presented the Multi-Features Siamese Tracker (MFST), a new tracking algorithm that exploits diverse feature hierarchies within the Siamese framework. We utilize features from different hierarchical levels and from different models using three combination strategies. Based on the feature combination, different abstraction levels of the target are encoded into a fused feature representation. Moreover, the tracker greatly benefits from the new feature representation due to our Squeeze-and-Excitation mechanism applied to different channels to recalibrate features. As a result, MFST achieved strong performance with respect to several state-of-the-art trackers.

\section*{Acknowledgments}
This work was supported by The Fonds de recherche du Québec - Nature et technologies (FRQNT) and Mitacs. We also thank Nvidia for providing us with the Nvidia TITAN Xp GPU.






%

\end{document}